\documentclass[letterpaper, 10 pt, conference]{ieeeconf}  % Comment this line out if you need a4paper

\IEEEoverridecommandlockouts                              % This command is only needed if 
\usepackage{graphics} % for pdf, bitmapped graphics files
\usepackage{epsfig} % for postscript graphics files
\usepackage{mathptmx} % assumes new font selection scheme installed
\usepackage{times} % assumes new font selection scheme installed
\usepackage{amsmath} % assumes amsmath package installed
\usepackage{amssymb}  % assumes amsmath package installed
\usepackage{algorithm}
\usepackage{algorithmic}
\usepackage{color}
\usepackage{booktabs}

\title{\LARGE \bf Trajectory-Safe Orienteering for Human-Robot Shared
  Environments 
}

\author{Songqun Gao$^{1,3}$, Elena Basei$^{2,3}$, Marco Roveri$^{2,3}$, Luigi Palopoli$^{2,3}$, Daniele Fontanelli$^{1,3}$ \thanks{This work is 
    supported by the EU project Magician (Grant Agreement n. 101120731).}  \thanks{$^{1}$ Department of Industrial Engineering,
    Università di Trento, Trento, Italy.}  \thanks{$^{2}$
    Department of Information Engineering and Computer Science,
    Università di Trento, Trento, Italy.}  \thanks{$^{3}$
    Interdepartmental Robotics Labs (IDRA), University of Trento,
    Trento, Italy {\tt\small \{name.surname\}@unitn.it}}
}

\begin{document}

\maketitle
\thispagestyle{empty}
\pagestyle{empty}

%%%%%%%%%%%%%%%%%%%%%%%%%%%%%%%%%%%%%%%%%%%%%%%%%%%%%%%%%%%%%%%%%%%%%%%%%%%%%%%%
\begin{abstract}
Orienteering problem (OP) has wide real-world applications and also great potential in human-robot collaboration. However, existing approaches struggle to simultaneously ensure safe and feasible trajectories while achieving high-quality task execution in shared workspaces. 
To this end, this work studies the OP with time windows and variable profits (OPTWVP). A two-stage DEcoupled discrete-Continuous Optimization with Service-time-guided Trajectory (DeCoST) approach is proposed to effectively solve OPTWVP in shared spaces. Meanwhile, the safety-aware time windows of nodes and the discretized workspace are introduced to ensure collision-free trajectories between the end effector and the human. Preliminary results validate the effectiveness of DeCoST in generating collision-free trajectory plans while preserving the quality of orienteering tasks.  
\end{abstract}

%%%%%%%%%%%%%%%%%%%%%%%%%%%%%%%%%%%%%%%%%%%%%%%%%%%%%%%%%%%%%%%%%%%%%%%%%%%%%%%%

\section{Introduction}
The Orienteering Problem (OP) is a classic combinatorial optimization problem with broad impact in factory scheduling, logistics, and robot planning. Recent progress on OP and its variants has largely come from two directions: (i) metaheuristics \cite{ils} that orchestrate multi-stage search procedures tailored to specific COP variants, and (ii) neural combinatorial optimization (NCO) methods \cite{gfacs,pomo} that leverage graph neural networks (GNN), Transformers, and other architectures to improve solution quality. However, these methods struggle to ensure that the solution remains feasible and collision-free when multiple agents share the same workspace in real applications. Task and motion planning (TAMP) \cite{tamp} can simultaneously consider the motion feasibility of a robot during task execution; it can also dynamically adjust the task plan when a task fails. However, TAMP methods still struggle to guarantee efficient completion of assigned tasks. 

In our work, we consider scenarios where robots share workspaces with humans: assuming the human trajectory is globally known, the robot is required to avoid the human while adjusting its plan without compromising the orientation task score. The problem is formulated as OP with time windows and variable profits (OPTWVP), and a learning-based approach, namely, DEcoupled discrete-Continuous Optimization with Service-time-guided Trajectory (DeCoST), is proposed to effectively solve OPTWVP in shared workspaces.

Specifically, the contribution of this work is given by:
\begin{itemize}
    \item DeCoST is proposed to decouple the discrete and continuous decision (service times) variables, which improves the inference speed and quality of the OPTWVP solution.
    \item The safety-aware time windows of nodes and the discretized workspace are introduced to ensure collision-free trajectories between the end effector and the human. 
    \item DeCoST is validated through preliminary results, demonstrating its ability to generate collision-free trajectory plans while preserving the task quality.   
\end{itemize}

\section{Methodology}
\subsection{Reinforcement Learning-based Orienteering Solver}
In this work, DeCoST is proposed to effectively decouple the discrete
and continuous decision variables in the OPTWVP problem, while
enabling efficient and learnable coordination between them. In the
first stage, a parallel decoding structure is employed to predict
the path and the initial service time allocation. The second stage
optimizes the service times through linear programming (LP)
formulation and provides a long-horizon learning of service time
estimation. We rigorously prove the global optimality of the
second-stage solution.  Experiments on OPTWVP instances demonstrate
that DeCoST outperforms both state-of-the-art constructive solvers
and the latest meta-heuristic algorithms in terms of solution
quality and computational efficiency, achieving up to 6.6x inference
speedup. Moreover, the proposed framework is compatible with various
constructive solvers and consistently enhances the solution quality
for OPTWVP. 

% \subsection{RL Decoder}
% \begin{algorithm}
% \caption{Selected node generation}
% \begin{algorithmic}[1]  % [1] 表示行号从 1 开始
% \IF{no target}
%     \STATE target $\leftarrow$ $Profitable\_Node\_Decoder()$
% \ENDIF
% % ======================
% \IF{target\_avail()}
%     \STATE selected $\leftarrow$ target
%     \STATE target $\leftarrow -1$ 
% \ELSE
%     \STATE selected $\leftarrow$ $Plan\_Decoder()$
% \ENDIF
% \STATE service\_time $\leftarrow$ $Service\_Time\_Decoder$
% \RETURN selected, service\_time
% \end{algorithmic}
% \end{algorithm}
\subsection{Collision Avoidance based on Time Windows}
The next step of this work is to integrate dynamic obstacle information into the graph representation. In this way, DeCoST is able to simultaneously search for an optimal orienteering plan and ensure that the resulting path is executable by the manipulator.

Assume the human trajectory is globally known during the task execution period. We evaluate the potential collisions between the human trajectory and the nodes in the graph. For any time interval during which a collision with the human trajectory occurs, the corresponding time window of the affected node is considered closed. Through this method, the information of dynamic obstacles in the workspace can be mapped onto the time window constraints of the nodes, thereby ensuring collision avoidance during the solution of the orienteering problem.

\subsection{Trajectory Generation}
In addition, to provide an executable solution for the manipulator, the two-dimensional workspace is discretized into M routing nodes, which are uniformly distributed over the workspace with a fixed resolution. These nodes are further embedded into the graph representation, and the graph nodes can be categorized into two types: routing nodes and profitable nodes. The routing node is an intermediate node and has no profit, $p=0$, while the profitable node is the place of interest relative to the task with $p>0$. 

During the solution phase, the agent reaches corresponding profitable nodes through routing nodes. Since the time windows of each node incorporate information about dynamic obstacles in the workspace, arriving at the target node during an open time window ensures the generated trajectory avoids collisions with moving objects.

Based on the waypoints generated by DeCoST, we collect intermediate nodes between different profitable nodes for trajectory generation. Based on these nodes, a motion plan for the robot is then generated to execute.

\section{Results and Discussion}
\subsection{Performance Evaluation of DeCoST}
To comprehensively evaluate the performance of the DeCoST framework on the extended OPTWVP problem, we compare it against several representative baselines, with results summarized in Table~\ref{tab:comparisonn500}. 
The results show that DeCoST consistently achieves strong performance
in all settings. It outperforms other heuristic and NCO methods in terms of solution quality (Score and Gap), while maintaining high
computational efficiency. 
Although runtime rises, the substantial increase in solution quality demonstrates that precise optimization of service time is crucial for achieving a better solution. % However, improvement in the quality of the solution underscores the value of precise optimization of service time.

\begin{table}[h]
\small
\centering
\caption{ Performance on OPTWVP with large-scale configuration
  (number of nodes = 500).  \textbf{Bold} values indicate the best
  result.}
% {\renewcommand{\arraystretch}{0.9}
\begin{tabular}{l|ccc}
\textbf{Method} & \textbf{Score} $\uparrow$ & \textbf{Gap} $\downarrow$ & \textbf{Runtime (ms)} $\downarrow$ \\
\hline
Branch \& Cut & 82.3 & 0.00\% & 68400 \\
ILS \cite{ils} & 78.2 & 4.98\% & 8803 \\
GFACS (Greedy) \cite{gfacs} & 67.4 & 18.1\% & 112 \\
GFACS & 73.1 & 11.3\% & 9420 \\
POMO \cite{pomo} & 58.6 & 28.8\% & 747 \\
% POMO + Edge Info & 58.9 & 28.4\% & 1344 \\
% POMO + Edge Info + STO & 77.6 & 5.77\% & 1367 \\
\textbf{DeCoST (Ours)} & \textbf{79.6} & \textbf{3.31\%} & 1329 \\
\end{tabular}
% }
\label{tab:comparisonn500}
\end{table}

\subsection{Performance Evaluation on Trajectory Generation}
We conducted a comparative experiment between our DeCoST and Branch \& Cut from a commercial solver, Gurobi. We used Gurobi to obtain the optimal score of the original OPTWVP (without workspace discretization, i.e., without routing nodes). As shown in Table~\ref{tab:comparison_76}, the gap between the two solutions is only 3.72\%, indicating that despite our approach considering workspace discretization to generate collision-free paths, it does not compromise the orienteering problem's score excessively. Additionally, our method is approximately 12 times faster than Gurobi, demonstrating its efficiency.

\begin{figure}[t]
    \centering
    \includegraphics[scale=0.28]{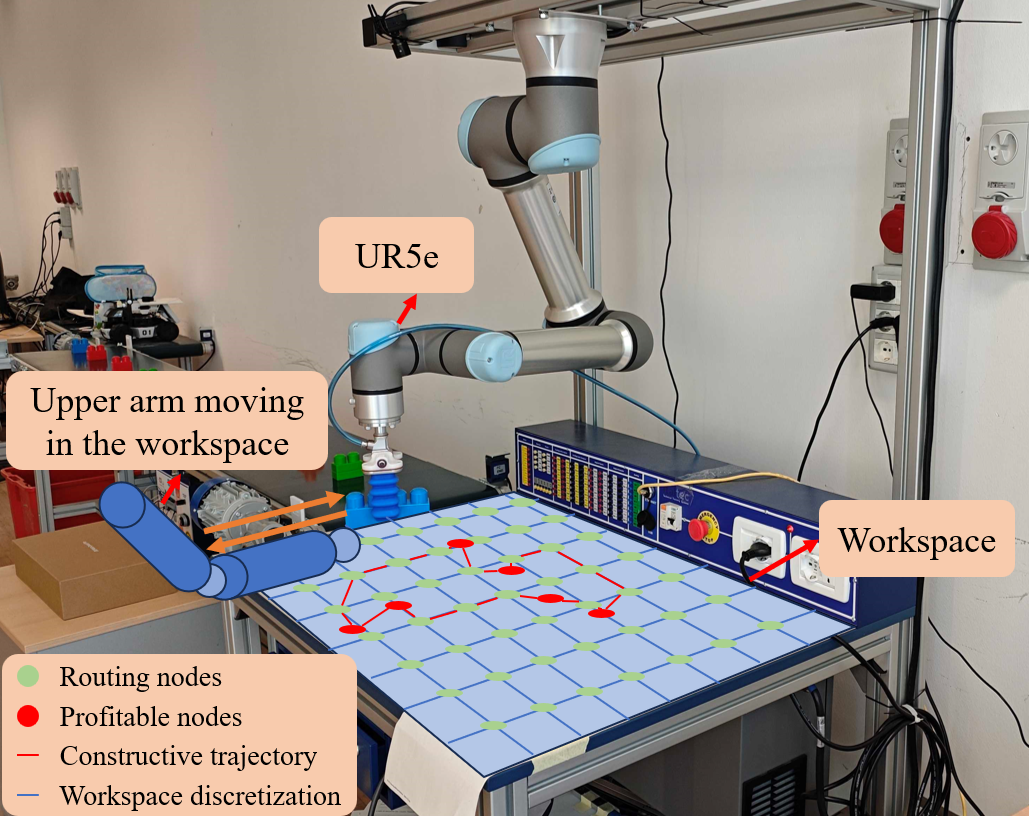}
    \caption{Simulation setup. We consider human and robot sharing the same workspace, and our proposed approach aims to find both an orienteering solution and a trajectory that is collision-free between the end effector and the human.}
    \label{fig:simulation_setup}
\end{figure}

Meanwhile, we conducted preliminary simulations of our proposed approach. In the setup, the moving obstacle is considered as an upper arm of a human, and humans and robots share the same workspace. The human is assumed to stand in front of the table and extend their upper arm into the workspace. Simulation setup is shown in Fig.~\ref{fig:simulation_setup}. Simulations show that our method allows the end-effector to avoid human motion in the workspace while still achieving a high orienteering score.

% 仿真显示我们的方法能够使end-effector 避开 human motion in the workspace。同时，依然取得高分数of in orienteering。该方法有希望能够在不同points of interest之间进行高效路由和路径规划，从而提高robot planning under human-robot shared workspace下的效率和适应性。
Simulation results show that our method enables the end effector to avoid human motion in the shared workspace while still maintaining high orienteering scores. This indicates its potential to efficiently route between different points of interest and to enhance the efficiency and adaptability of robot planning in human-robot shared workspaces.
\begin{table}[h]
\small
\centering
\caption{ Performance evaluation against the original OPTWVP.  \textbf{Bold} values indicate the best
  result.}
\begin{tabular}{l|ccc}
\textbf{Method} & \textbf{Score} $\uparrow$ & \textbf{Gap} $\downarrow$ & \textbf{Runtime (ms)} $\downarrow$ \\
\hline
Branch \& Cut & 31.07 & 0.00\% & 1010 \\
DeCoST & 29.91 & 3.72\% & 79.46
\end{tabular}
\label{tab:comparison_76}
\end{table}

% 该方法有希望能够在human-robot collaboration中提供可行的路径


\end{document}